\documentclass[11pt]{article}

\usepackage[final]{acl}

\usepackage{times}
\usepackage{latexsym}
\usepackage{todonotes}
\usepackage{tabularx}
\usepackage{longtable}
\usepackage{array}
\usepackage[table,xcdraw]{xcolor}
\usepackage{booktabs}
\usepackage{multirow}
\usepackage{graphicx} 
\usepackage{booktabs} 
\usepackage{arydshln}
\usepackage{url}

\usepackage[T1]{fontenc}

\usepackage[utf8]{inputenc}

\usepackage{microtype}
\usepackage{makecell}
\usepackage{amsmath}
\usepackage{inconsolata}

\usepackage{graphicx}
\usepackage{adjustbox}
\title{MATEO: A Multimodal Benchmark for  \\ Temporal Reasoning and Planning in LVLMs}

\author{ Gabriel Roccabruna\textsuperscript{\ $\dagger$\ }\thanks{Work done while at University of Trento, prior to joining Amazon.}, Olha Khomyn\thanks{Equal contribution.}, Giuseppe Riccardi  \\
        Signals and Interactive Systems Lab, \\ University of Trento, Italy \\ 
        \texttt{giuseppe.riccardi@unitn.it} }

\begin{document}

\maketitle
\begin{abstract}

AI agents need to plan to achieve complex goals that involve orchestrating perception, sub-goal decomposition, and execution.  These plans consist of ordered steps structured according to a Temporal Execution Order (TEO, a directed acyclic graph that ensures each step executes only after its preconditions are satisfied.  Existing research on foundational models’ understanding of temporal execution is limited to automatically derived annotations, approximations of the TEO as a linear chain, or text-only inputs. To address this gap, we introduce MATEO (MultimodAl Temporal Execution Order), a benchmark designed to assess and improve the temporal reasoning abilities of Large Vision Language Models (LVLMs) required for real-world planning. We acquire a high-quality professional multimodal recipe corpus, authored through a standardized editorial process that decomposes instructions into discrete steps, each paired with corresponding images.  We collect TEO annotations as graphs by designing and using a scalable crowdsourcing pipeline. Using MATEO, we evaluate six state-of-the-art LVLMs across model scales, varying language context, multimodal input structure, and fine-tuning strategies. 




\end{abstract}

\section{Introduction}
 
\begin{figure}[h!]
    \centering
    \includegraphics[width=0.37\textwidth, trim={1.2cm 1.1cm 1.2cm 1.1cm}, clip=true]{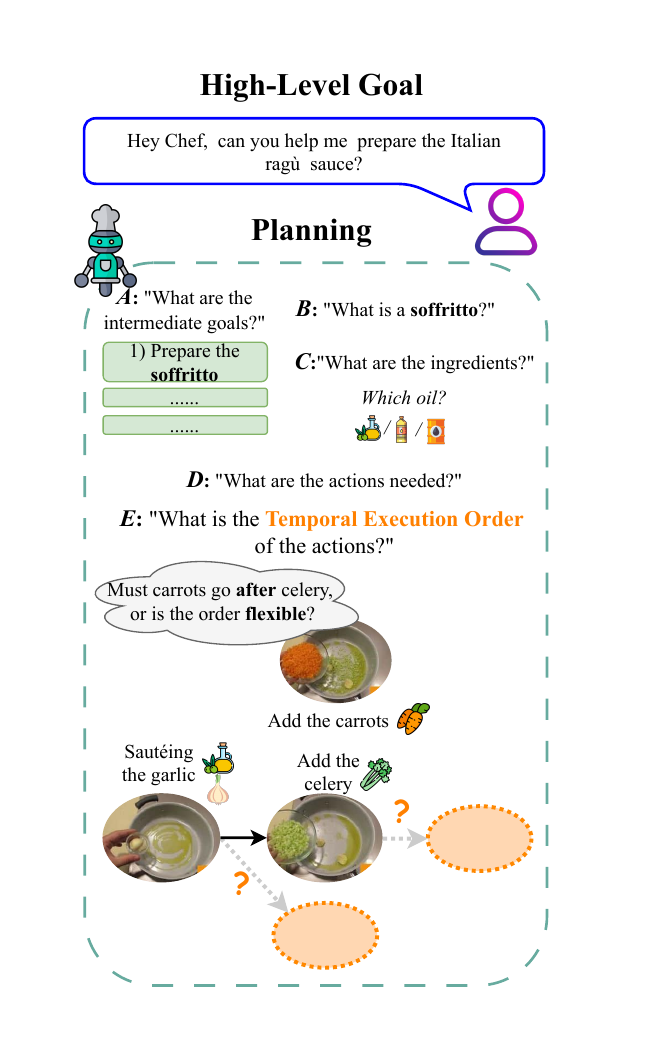}
    \caption{Example of an AI agent’s planning process and inherent uncertainties for a natural-language goal. Questions\textit{ \textbf{A–D}} decompose the high-level goal into executable actions, while \textbf{\textit{E}} infers their Temporal Execution Order (TEO) as a directed acyclic graph. Each step introduces uncertainty, producing multiple possible paths, some correct, others wrong. }
    \label{fig:task-figure}
\end{figure}
The era of autonomous agents with natural language interfaces has recently gained attention from both academia and industry.  AI agents have been successfully adopted across many fields, including finance \cite{fatemi2024enhancing, zhang2024multimodal}, software development \cite{suri2023software, mo2025interactive}, and networking \cite{zhang2023ai}. Agents are autonomous systems that can accomplish goals with little or no human supervision, a concept that dates back to the earliest days of artificial intelligence \cite{weizenbaum1966eliza, wooldridge1995intelligent}. Several frameworks have been proposed to describe the architecture of these systems, including the BDI (belief-desire-intention) model \cite{rao1995bdi} or, more recently, the perceive-reason-act-learn paradigm in agentic AI \cite{raheem2025agentic}. Despite differences in terminologies and definitions, these frameworks generally decompose the agent behavior into environmental sensing, reasoning and understanding, planning, and action execution.

Planning is one of the most complex components, as it requires mapping high-level goals to executable action sequences. Its complexity increases when the input is natural language processing alongside other sources, such as images or sensor signals. Classical symbolic planners, such as STRIPS \cite{fikes1971strips}, represent as logic conditions to be satisfied through actions defined by explicit pre- and post-conditions. Although these methods offer strong guarantees of soundness and completeness and are explainable by design, they depend on hand-crafted domain models and thus scale poorly to dynamic environments \cite{acharya2025agentic}. 

Foundation language models offer a promising alternative, as they can directly manipulate goals and actions expressed in natural language without requiring a formal symbolic description, making them flexible to adapt to dynamic environments \cite{raheem2025agentic}. To generate an effective plan, a language model first needs to parse the goal and available actions, while addressing the inherent ambiguity of natural language and relative uncertainties, by following a process similar to that depicted in Figure \ref{fig:task-figure}. Specifically, it then needs to reason over this inferred information, together with parametric knowledge or retrieved external data, to construct a sequence of actions that achieves the desired goals.

Current research on Large Language Models (LLMs) and Large Vision-Language Models (LVLMs) as planners has shown serious limitations when dealing with complex tasks such as travel, flight, and calendar planning \cite{10.5555/3692070.3694316, 10.1145/3746027.3755790, zheng2024}. Nonetheless, most assessments of these focus on plan-level metrics, such as the success rate, providing little insight into the fine-grained sources of errors. While planning is generally considered a PSPACE-complete problem \cite{bylander1994computational}, understanding whether these models can reliably infer the preconditions and effects of actions, which is a fundamental ability for any planning system, might help improve their performance and deepen our understanding of their actual reasoning abilities.

A possible way to investigate these abilities is to evaluate the models' performance on the Temporal Execution Order (TEO) task \cite{chambers2013navytime, derczynski2017automatically, vashishtha2020temporal}. In this, given two steps representing two actions, the model has to infer their temporal relation, such as which action should occur first or second, or if the actions are order-independent. Successful resolution of this requires reasoning over the actions’ implicit pre- and post-conditions. Procedural texts, such as recipes or instruction books, provide an ecologically valid and scalable source of such instances, as they explicitly encode action sequences as plans expressed through stepwise natural-language instructions. However, existing benchmarks rely on text-only procedural text \cite{mori-etal-2014-flow, lal-etal-2024-cat}, automatically derived annotations, or approximate TEO as a strictly linear chain \cite{wu-etal-2022-understanding, lu-etal-2024-multimodal,qiu2025}. This limits the interpretation of the model's actual performance on the TEO and the insights into whether LVLMs can function as \textit{world models}, i.e., systems capable of grounding plans on world observations \cite{chen2025planning}.

In this work, we introduce MATEO (\textbf{M}ultimod\textbf{A}l \textbf{T}emporal \textbf{E}xecution \textbf{O}rder), a publicly available benchmark designed for evaluating and improving the multimodal temporal reasoning abilities of LVLMs. MATEO assesses a model’s capability to determine the Temporal Execution Order (TEO) of a sequence of multimodal actions, a fundamental building block for planning in real-world contexts. The benchmark comprises 300 high-quality professional recipes from a well-known Italian recipe website\footnote{\url{https://www.giallozafferano.it/}}. Each recipe is organised as a sequence of steps, with every step containing both a textual description and an image illustrating the action or its outcome, ensuring quasi-perfect semantic alignment between text and visual content. Different from most existing human-annotated recipe corpora \cite{mori-etal-2014-flow, yamakata-etal-2020-english, Pan2020MultimodalCW}, which rely on internal annotators for TEO labelling, we employ crowdsourcing, providing a scalable and reproducible annotation methodology. The annotation process has produced a Directed Acyclic Graph (DAG) for each recipe, capturing the pre- and post-condition dependencies (\textit{edges}) of each action (\textit{nodes}). Using this benchmark, we evaluate six open and closed-sourced LVLMs with various promoting strategies, including in-context learning \cite{brown2020language}, chain of thought \cite{wei2022chain}, and self-reflection \cite{madaan2023self}, and fine-tuning approaches. Our findings reveal that while some models achieve state-of-the-art results with multimodal inputs, the majority still struggle to effectively leverage both modalities. Furthermore, even the best performing models only achieve 0.69 accuracy, highlighting suboptimal capabilities in the TEO task and motivating future adoption of MATEO for developing novel methods to enhance temporal reasoning and real-world planning.


In summary, the main contributions of this paper are:
\begin{itemize}
    \item Release MATEO, a publicly available\footnote{GitHub: \url{https://github.com/sislab-unitn/MATEO}} multimodal benchmark for evaluating and advancing the temporal reasoning abilities of LVLMs through the Temporal Execution Order (TEO) task;
    \item A systematic evaluation of six open- and closed-source LVLMs using in-context learning (ICL), chain-of-thought prompting (CoT), self-reflection, and fine-tuning approaches while varying the input modalities;
    \item A set of novel guidelines for extending the benchmark to new recipes, other languages, and domains, ensuring scalability and reproducibility.
\end{itemize}

\section{Literature Review}

\textbf{Planning}  The task of planning is to find a sequence of actions that, when executed, transitions a system from an initial state to a desired goal state \cite{10.5555/3666122.3669442}. Recent work has investigated LLMs' planning abilities by evaluating them on classical planning problems, such as Blocks World \cite{gupta1991complexity}, showing their limitations in symbolic planning \cite{10.5555/3666122.3669442, 10.5555/3692070.3692991}. However, these planning problems might be overspecific and formalized, impeding the model from leveraging any parametric knowledge, and are described with hand-crafted actions and states, impacting the flexibility to adapt to dynamic environments. In this regard, several benchmarks in which initial and goal states are written in natural language have been proposed. They cover various domains such as the planning of trips, meetings, and calendars \cite{zheng2024, 10.5555/3692070.3694316}. Nonetheless, these are only for sequential planning, i.e., without accounting for possible execution of simultaneous actions, and most of the proposed assessments are based on plan-level metrics. A few studies represent the plan as a graph, but they rely on machine-generated annotations \cite{10.5555/3692070.3693283}. Although multimodal planning is a relatively new topic, a few benchmarks for investigating the planning abilities of LVLMs have been proposed, such as MPCC \cite{10.1145/3746027.3755790}, EgoPlan-Bench2 \cite{qiu2025}, ALFWorld \cite{ijcai2024p15}, and procedural text corpora \citet{wu-etal-2022-understanding, lu-etal-2024-multimodal}. However,  all of these are limited to a linear chain of action due to goals with low uncertainty or a lack of an ad-hoc annotation.

\textbf{TEO in Recipes} Recipes constitute a challenging multimodal planning domain and a rich procedural-text genre for the TEO task, as their steps often form parallelizable sub-plans with partial ordering constraints. Among the earliest efforts to model this planning complexity are the Japanese Recipe Flow Graph \cite{mori-etal-2014-flow} and the English Recipe Flow Graph \cite{yamakata-etal-2020-english} corpora, which model directly the dependencies and effects of each action expressed by verbs, similarly to a dependency parser. Step-level dependencies have been extracted from the English Recipe Flow Graph to create the CaT-Bench corpus \cite{lal-etal-2024-cat}. Although a multimodal recipe corpus following a similar logic has been proposed \cite{Pan2020MultimodalCW}, no public multimodal DAG corpus is currently available. Existing resources rely on non-curated web/video data, leading to inconsistent formats\footnote{e.g., variable image counts per step, mixed unit systems.}, and their annotations are produced by non-crowdworkers, limiting scalability and reproducibility.

\section{Methodology}
\label{sec:methodology}
To evaluate the reasoning abilities required for planning in LLMs and LVLMs, we introduce the Temporal Execution Order (TEO) task. We formulate TEO as a three-way classification problem over the temporal execution relations \textit{before} and \textit{after}, and \textit{independent}. These relations are semantically aligned with the \textit{before}, \textit{after}, and \textit{simultaneous} temporal relations defined in Allen's interval algebra \cite{allen1983maintaining}. However, unlike Allen's algebraic relations, which are derived from comparisons of interval start and end points, our formulation infers temporal execution order from logical dependencies rather than explicit time boundaries. In particular, if action B requires a condition that is produced by the post-condition (aka effect) of action A, then A must be executed \textit{before} B (equivalently, B must occur \textit{after} A). Conversely, if A and B do not depend on each other’s post-conditions, including indirect or transitive dependencies (e.g., $A \rightarrow C \rightarrow B$, where B requires C and C requires A), then the two actions are considered independent, meaning they can be executed in any order. Given a set of action instances $\mathcal{A}$ and their dependency relations  $\mathcal{E}$ , TEO can be represented as the graph $G= (\mathcal{A}, \mathcal{E})$, where $\mathcal{A}$ are the nodes, and $\mathcal{E}$ are the edges. Specifically, this graph is a directed acyclic graph (DAG) because the dependencies have an orientation and any cycle would generate a paradox, making the plan infeasible (e.g., an action dependent on its own output). 


Similar to \citet{roccabruna2024will} and \citet{lal-etal-2024-cat}, for each class, we pose a corresponding ternary (Yes/No/I don't know) question, as presented in Table \ref{tab:teo_questions}.

\begin{table}[h!]
\centering
\begin{adjustbox}{max width=\linewidth}
\begin{tabular}{l|p{6cm}}
\textbf{TEO class} & \textbf{Question} \\ \toprule 
Before       & Must Step A be executed \textbf{before} Step B?  \\
\midrule
After         & Must Step A be executed \textbf{after} Step B?  \\
\midrule
Independent        & Can Step A and Step B be executed in \textbf{parallel}?  \\
\end{tabular}
\end{adjustbox}
\caption{Temporal Execution Order (TEO) classes and their corresponding ternary questions.}
\label{tab:teo_questions}
\end{table}

To investigate whether the models can answer these questions, we formalise the following input sequence. Since our corpus consists of recipes, we refer to actions as steps, following the natural structure of a recipe. In our corpus, a step is composed of an image $I$ and a textual part $T$, formally $S_j = I_j \oplus T_j$. Thus, given two steps, A and B, the model is tasked to answer the three TEO questions of Table \ref{tab:teo_questions}, denoted as $Q_i$,  using the following input sequence:
\[ C \oplus S_A \oplus S_B \oplus Q_1 \oplus Q_2 \oplus Q_3\ \]
where $C$ is the grounding context, $\oplus$ denotes concatenation with a newline ("\textbackslash n"). Marker tokens (``Step A picture:'', ``Step A description:'', etc.) precede each image and text component to further guide the model.

The TEO class is then retrieved by parsing the model's output by assigning class $i \in \{\textit{before}, \textit{after}, \textit{independent}\}$, if the model answers \textit{``Yes''} to the corresponding $Q_i$ TEO question and \textit{``No''} to both other $Q_j$ for $j \neq i$. All the other combinations, such as answering \textit{``Yes''} to multiple questions or responding \textit{``I don’t know''}, are assigned to the class \textit{Other}. This captures inconsistent behavior, such as simultaneously affirming the \textit{before}  and \textit{after} questions, as well as uncertainty. 

In Chain of Thought experiments \cite{wei2022chain}, the model is prompted to directly predict one of the three TEO classes. The reason for this was to simplify and shorten the demonstrations provided in the prompt.  


\section{The Benchmark}
Our requirements for building MATEO were to identify a form of procedural text in which the textual and visual components are maximally semantically aligned, enabling reliable multimodal reasoning over execution order. We have found that these criteria are met by the professionally edited recipes on GialloZafferano, a well-known Italian-language recipe website. These recipes offer several advantages: i) they follow an editorially-curated narrative and structural format; ii) the images are professionally produced, visually consistent, and captured from a near-uniform top-down perspective, minimizing viewpoint variability; and iii) each recipe is explicitly segmented into an ordered sequence of steps. Each step typically describes a single executable action, and the accompanying image depicts either the action's outcome or the action in progress. Together, these properties reduce multimodal noise sources and enable cleaner corpus utilization \footnote{Because GialloZafferano's recipes are copyright-protected and not publicly licensed for AI benchmarking, the publisher agreed to support our research initiative by providing a subset of approximately 300 recipes for release to the research community. The recipes were randomly selected via stratified sampling over dish category (e.g., appetizers, main courses, desserts) to ensure balanced coverage across recipe types.}.

To build MATEO, we employed human annotators recruited through Prolific\footnote{\url{https://www.prolific.com/}}, a crowdsourcing platform. To balance crowdworker cognitive load, we divided the annotation process into batches of five recipes, with an average completion time of approximately 50 minutes per batch. We considered only native Italian speakers, given that the recipes are in Italian. We set the compensation to \textsterling 9.60 per hour, which falls within the platform's recommended range.

The annotators were provided with a list of steps in the order of the original recipe. They were tasked with linking these recipe steps into a directed acyclic graph by following the TEO relations by an ad-hoc UI (Figure \ref{fig:dag_ex} and Figure \ref{fig:annotation_platform} in the Appendix, respectively). To explain the task concisely and effectively, we provided guidelines, a short demo video, and multimodal instructions (textual and visual), along with explanatory examples.  The annotation platform and the guidelines\footnote{They will also be shared on GitHub along with the code used to conduct our experiments.} are shown in Appendix \ref{sec:materials}.

Additionally, to ensure data quality, annotators were required to pass a qualification test before starting the actual task. Only those who reached at least 65\% accuracy on two test recipes were allowed to continue with the actual annotation work. The accuracy is computed as if it were a multi-label classification task, where the examples are the nodes and the labels are the outgoing nodes\footnote{To give a better visualization the data structure is $n_1: [ n_2, n_3], .., n_2: [..]$, where $n_2$ and $n_3$ are the outgoing nodes and the ``labels'' of $n_1$.}. The threshold corresponds to the 75th percentile of the score distribution computed from testing 25 crowdworkers who were not involved in the actual annotation. 

In total, 54 annotators\footnote{Some of the annotators participated multiple times. This is because we experienced a shortage of annotators willing to perform our task, probably due to a high rejection rate.} annotated 315 recipes, collected in 63 batches. We computed the annotation agreement using MASI distance metrics \cite{passonneau-2006-measuring}, which are specifically designed for measuring agreement on multi-labelling classification tasks. The annotators reached an agreement score of 0.85 (substantial agreement)\footnote{The agreement was computed on the examples used for qualifying the annotators. Although this may give an overestimation of the actual agreement, it was the best trade-off as we could not afford to have an overlap on the annotation due to the high amount of time needed to complete a batch.}. Moreover, we have manually checked the annotations for specific error patterns, such as graphs with isolated steps or graphs that were fully linear. At the end of this process, we discarded 15 recipes due to poor annotation quality. 

The resulting corpus contains 300 Italian recipes. Preliminary experiments show that some LVLMs perform worse on this task in Italian than in English, probably because these models are trained primarily on English \cite{bai2025qwen25vltechnicalreport}.  For this, we translate the recipe steps from Italian to English using Llama-3-8B \cite{grattafiori2024llama3herdmodels}, given its relatively small model size and acceptable performance. Before translating the full corpus, we manually verified a subset of the translations and found them to be of high quality. All the following experiments are conducted on the English translations. The dataset is split into training, validation, and test sets using stratified sampling by number of steps, with a 70\%/10\%/20\% split. The statistics of the splits are reported in Table \ref{tab:dataset_stats}.

\begin{table}[t]
\centering
\begin{adjustbox}{max width=\linewidth}
\renewcommand{\arraystretch}{1.2} %
\begin{tabular}{l|c|c|c}
& \textbf{Train} & \textbf{Valid} & \textbf{Test} \\ \toprule 
\# Recipes        & 210 & 30 & 60 \\
\# Steps        & 3273 & 477 & 936 \\
AVG Steps         & 15.6 $\pm$ 3.9 & 15.9 $\pm$ 3.7 & 15.6 $\pm$ 3.8 \\
\makecell[l]{Branching\\Factor} & 1.12 $\pm$ 0.43 & 1.16 $\pm$ 0.42 & 1.11 $\pm$ 0.38 \\
\end{tabular}
\end{adjustbox}
\caption{Splits of the annotated corpus used for training and evaluating LVLMs’ capabilities. The table reports the number of samples, the average number of steps, and the branching factor.}
\label{tab:dataset_stats}
\end{table}

\section{Experimental settings}
\subsection{Models}
We select six LVLMs that vary in model size, with selection primarily based on their support for sequences of images in the input. Our evaluation includes both open-source, namely, Qwen2.5-VL-7B, Qwen2.5-VL-72B \cite{bai2025qwen25vltechnicalreport}, LLaVA-OneVision-7B \cite{li2024llavaonevisioneasyvisualtask}, InternVL3.5-8B, InternVL3.5-38B \cite{wang2025internvl}, and the closed-source model GPT-5.1\footnote{https://platform.openai.com/docs/models/gpt-5.1}. We could evaluate GPT-5.1 on only 20\% of the test set, selected via stratified sampling, due to budget constraints.

\subsection{Dataset}
Given the annotated DAG for each recipe, we infer the TEO classes. In a real-world setting, determining temporal relations would require comparing all pairs of steps, resulting in $n^2$ comparisons (876,096 pairs in our test set). To make evaluation tractable, we restrict comparisons only to annotated dependencies: step A must be executed \textit{before} step B if there is a directed edge from A to B. The class \textit{after} is the inverse of \textit{before}, therefore, we have obtained the ground truth examples by swapping the steps A and B. Steps A and B are \textit{independent} if there is no directed path from A to B or from B to A. The number of resulting relations is 1573 for the \textit{independent} class, 993 for \textit{before}, and 933 for the \textit{after} classes. 

We evaluate all models on two versions of the dataset. The first preserves the original annotation order, yielding \textit{before} and \textit{independent} labels. The second reverses each step pair, mapping \textit{before} to \textit{after}, while \textit{independent} labels remain unchanged. With this, we can assess whether the model actually understands the relations among the steps, or whether it relies on biases learnt during pre-training. 


\subsection{Prompt settings}
We evaluate the planning abilities of LVLMs on the TEO task under different modalities and grounding contexts. Specifically, we compare text-only, image-only, and multimodal inputs, as described in Section \ref{sec:methodology}. Below, we describe the grounding contexts we used to evaluate the models. The example for each prompt template can be found in Table \ref{table:prompt_schema} in the Appendix \ref{sec:materials}.

\textbf{Baseline} The model is only conditioned on two recipe steps and the three TEO questions.

\textbf{Instructions} We provide the model with a set of rules that explain the meaning of the TEO classes and instruct the model to respond to each question. The rules are general for the TEO task and are not specific to the cooking domain.

\textbf{In-Context Learning (ICL)} The model is conditioned on the instructions described previously and three in-context examples for each TEO class. Each example consists of a pair of steps, the TEO questions, and the correct answers, accompanied by brief explanations to encourage the model to emulate the latent reasoning process. To limit context length and computation time, all in-context examples are text-only. We prepend a modality-specific instruction noting that examples contain only text descriptions, while the model input may also include images.

\textbf{Chain-of-Thought (CoT)} Correctly predicting the TEO class requires understanding whether the outcome of one step is a prerequisite for completing the other. In this experiment, we explicitly ask the model to identify pre- and post-conditions of each step and then reason about their dependency. After this reasoning process, the model is tasked to directly predict the TEO class.

\textbf{Self-Reflection} At the end of the CoT, the model is instructed to reflect on its answer by reviewing any inconsistencies or errors in the reasoning process, and then give the final answer.

\subsection{Fine-Tuning}
Based on the results of the prompting experiments, we have selected the best-performing small-scale model, InternVL3.5-8B, for fine-tuning. We fine-tune the model with LoRA \cite{hu2021loralowrankadaptationlarge} on image-only, text-only, and image-text modalities following the \textit{Instructions} prompt task. To prevent label imbalance, we swap only dependent step pairs to generate examples for the \textit{after} class. For independent step pairs, swapping does not change the label and would therefore introduce duplicate examples, potentially biasing the model toward the \textit{independent} class. This resulted in 3,499 examples for each of the \textit{before} and \textit{after} classes, and 4,969 for \textit{independent}. Examples of model inputs are provided in Table \ref{tab:fine_tuning_ex} in the Appendix  \ref{sec:hyperparameters} along with all the hyperparameters.

\section{Results and Error Analysis}

\begin{table*}[h!]
\centering
\begin{adjustbox}{max width=\textwidth}
\begin{tabular}{l|l|c|c|c|c|c}

\textbf{Model} & \textbf{Modality} & \textbf{Baseline} & \textbf{Instructions} & \textbf{ICL} &
\textbf{CoT} & \textbf{Self-Reflect.}  \\ 
\toprule

\multirow{3}{*}{Qwen2.5-VL-72b}
& Image-only & 0.08 & 0.55 & 0.58 & 0.38 & 0.34  \\
& Text-only & 0.14 & 0.53 & 0.57 & 0.50 & 0.50\\
& Image+Text & 0.17 & 0.61 & \textbf{0.68} & 0.59 & 0.54 \\
\hline

\multirow{3}{*}{InternVL3.5-38b}
& Image-only & 0.01 & 0.51 & 0.53 & 0.49 & 0.44 \\
& Text-only & 0.10 & 0.50 & \textbf{0.59} & 0.37 & 0.39 \\
& Image+Text & 0.09 & 0.51 & 0.49 & 0.50 & 0.47 \\
\hline

\multirow{3}{*}{InternVL3.5-8b}
& Image-only & 0.15 & \textbf{0.58} & 0.39 & 0.14 & 0.19 \\
& Text-only & 0.10 & 0.31 & 0.46 & 0.31 & 0.32 \\
& Image+Text & 0.10 & 0.34 & 0.32 & 0.33 & 0.24 \\
\hline

\multirow{3}{*}{Qwen2.5-VL-7b}
& Image-only & 0.01 & 0.05 & 0.04 & 0.09 & 0.05 \\
& Text-only & 0.01 & 0.11 & 0.08 & 0.15 & 0.09 \\
& Image+Text & 0.02 & 0.10 & 0.03 & 0.17 & \textbf{0.21} \\
\hline

\multirow{3}{*}{LLaVA-OneVision-7b}
& Image-only & 0.00 & 0.00 & 0.00 & 0.02 & 0.00 \\
& Text-only & 0.00 & 0.00 & 0.07 & \textbf{0.22} & 0.09 \\
& Image+Text & 0.00 & 0.03 & 0.04 & 0.09 & 0.00 \\
\hline

\multirow{3}{*}{GPT-5.1}
& Image-only & 0.13* & 0.54* & 0.63* & 0.71* & 0.71* \\
& Text-only & 0.07* & 0.27* & 0.27* & 0.62* & 0.62* \\
& Image+Text & 0.29* & 0.58* & 0.7* & 0.73* & \textbf{0.74*} \\
\hline
\hline

\multirow{3}{*}{InternVL3.5-8B\textsubscript{Fine-tuned}}
& Image-only & - & 0.68 & - & - & - \\
& Text-only & - & 0.61 & - & - & - \\
& Image+Text & - & \textbf{0.69} & - & - & - \\

\end{tabular}
\end{adjustbox}
\caption{Performance of LVLMs on the TEO task under different prompt strategies and fine-tuning results. We report accuracy by considering the consistency of predictions across the original and swapped orders of steps. The results in bold indicate the best results per model. *Results achieved on 20\% of the test set only.}
\label{tab:results}
\end{table*}

To evaluate models' performance in predicting temporal execution order, we consider both the original and the reverse order of each step pair. This setting reflects realistic scenarios in which the true execution order may be unknown. Accuracy is reported based on a consistency criterion: a prediction is considered correct only if the model remains consistent across both the original and swapped step orders. Specifically, for dependent pairs, the model must predict \textit{before} in the original order and \textit{after} when swapped. For independent pairs, the model must predict \textit{independent} in both configurations.

Table \ref{tab:results} reports accuracy scores across prompting strategies, and after fine-tuning. All results are obtained from a single evaluation using greedy decoding to ensure reproducibility. Starting with the prompting strategies, we observe that in the baseline setting, all models perform poorly, with near-zero accuracy, indicating limited zero-shot ability on the TEO task. Adding task instructions, in-context learning, or chain-of-thought improves performance for most models, but the effectiveness of each setting depends on the model; thus, there is no universal best strategy. Furthermore, results show that self-reflection sometimes leads to lower performance compared to CoT, suggesting a limitation of this prompting strategy for some models.

Regarding individual model performance, Qwen2.5-VL-72B achieves the highest overall accuracy among the open-source LVLMs. When using ICL with image and text inputs, it attains an accuracy of 0.68. Under the same prompting strategy, InternVL3.5-38B achieves its best performance using text-only inputs, reaching 0.59 accuracy. Among models with fewer than 10B parameters, Qwen2.5-VL-7B and LLaVA-OneVision-7B substantially underperform, with accuracy at or near zero in most settings. In contrast, InternVL3.5-8B achieves performance comparable to larger-scale models, reaching 0.58 accuracy when prompted with task instructions and image-only inputs. While the GPT-5.1 results are not directly comparable\footnote{Not comparable but indicative as Qwen2.5-VL-72B on the same partition with ICL Image+Text achieves  0.66 compared to 0.68 on the whole test set.} (we evaluate it only on 20\% of the test set), our findings suggest that it may perform better than the open-source models, achieving 0.74 accuracy with self-reflection and multimodal inputs. In contrast, some models fail to self-reflect or produce a final answer, with failure rates ranging from 2.71\% for Qwen2.5-VL-72B to 81.6\% for LLaVA-OneVision-7B, compared to just 0.03\% for GPT-5.1. When self-reflection is successful, models largely preserve their original predictions (98.5\% for GPT-5.1 and 97.0\% for Qwen2.5-VL-72B). Notably, GPT-5.1 makes only correct self-corrections (0.9\%), whereas Qwen2.5-VL-72B makes nearly equal numbers of correct and incorrect self-corrections (0.14\% and 0.13\%), suggesting a possible reason for GPT-5.1’s stronger gains under this strategy.

Although multimodal input leads Qwen2.5-VL-72b and GPT-5.1 to achieve state-of-the-art results, not all models show the same ability to leverage this modality. Indeed, other models struggle to incorporate multimodal information, and we observe a drop in performance compared to single modalities. A possible explanation could be that the scale of the language decoder, with small-scale models, fails to effectively attend to different modalities. 

\begin{table*}[h!]
\centering
\begin{adjustbox}{max width=\textwidth}

\begin{tabular}{l|l|c|c|c|c|c}
\textbf{Model} & \textbf{Class} & \textbf{Baseline} & \textbf{Instructions} & \textbf{ICL} &
\textbf{CoT} & \textbf{Self-Reflect.} \\ 
\toprule

\multirow{3}{*}{GPT-5.1}
& Before      & 0.8* & 0.82* & 0.79* & 0.78* & 0.79* \\
& Independent  & 0.39* & 0.79* & 0.85* & 0.87* & 0.87* \\
\cdashline{2-7}
& After        & 0.65* & 0.76* & 0.7* & 0.69* & 0.72* \\
\midrule

\multirow{3}{*}{Qwen2.5-VL-72b}
& Before      & 0.6 & 0.76 & 0.79 & 0.79 & 0.78 \\
& Independent  & 0.21 & 0.79 & 0.85 & 0.86 & 0.86 \\
\cdashline{2-7}
& After        & 0.5 & 0.71 & 0.71 & 0.42 & 0.43 \\
\midrule

\multirow{3}{*}{InternVL3.5-8b}
& Before      & 0.6 & 0.35 & 0.63 & 0.71 & 0.67 \\
& Independent  & 0.0 & 0.73 & 0.63 & 0.74 & 0.69 \\
\cdashline{2-7}
& After        & 0.33 & 0.51 & 0.47 & 0.18 & 0.16 \\

\end{tabular}
\end{adjustbox}
\caption{Per-class F1 scores for the TEO task. Scores above the dashed line are computed on the original step order, and those below on the swapped order. Models in zero-shot show bias toward \textit{before} and \textit{after} relations, lowering \textit{independent} performance. Prompting improves \textit{independent} and \textit{after} scores; however, \textit{after} remains challenging, suggesting difficulties in reasoning over reversed temporal order. *Results achieved on 20\% of the test set only.}
\label{tab:class_results}
\end{table*}

Regarding the fine-tuning experiments, InternVL3.5-8B achieves 0.69 of accuracy, which is the highest among other experiments with open-source models. When restricted to a single modality, InternVL3.5-8B also outperforms all other open-source models in that modality. Notably, image-only fine-tuning yields better performance than text-only fine-tuning. Some possible explanations are that the model cannot fully leverage textual input alone, or that images more precisely capture the underlying actions, whereas textual descriptions may contain superfluous information, such as references to other steps, or lack sufficient contextual information.

We analyze performance across individual TEO classes to identify the most challenging cases. Table \ref{tab:class_results} reports F1 scores for the \textit{before}, \textit{independent}, and \textit{after} classes. Because the ground-truth labels for the \textit{after} class are defined on the swapped step order, we compute F1 scores for this class on the swapped dataset. In contrast, scores for the \textit{before} and \textit{independent} classes are computed using the original step order. For clarity, we present the selection of three models, all under multimodal inputs; similar trends are observed across all models. Under the baseline zero-shot setting, models exhibit a bias toward \textit{before} and \textit{after} relations, lowering performance on the \textit{independent} class. Performance on the \textit{independent} class improves substantially with task instructions, ICL, and CoT prompting, while gains for the \textit{before} class are more moderate. For Qwen2.5-VL-72B, CoT yields relative improvements of 32\% and 310\% for the \textit{before} and \textit{independent} classes with respect to the baseline, respectively. Moreover, the models struggle in predicting the \textit{after} class, particularly under the baseline and CoT settings. This suggests a bias inherited from pre-training, where models have difficulty reasoning under reversed TEO. When conditioned on task instructions, Qwen2.5-VL-72B achieves an F1 score of 0.71 on the \textit{after} class, while InternVL3.5-8B reaches a maximum F1 of 0.51. 

Higher per-class scores compared to the consistency-based accuracy indicate that models struggle to produce consistent predictions across the original and swapped step orders. Larger-scale models, namely Qwen2.5-VL-72B and InternVL3.5-38B, exhibit greater consistency in their predictions. In particular, they produce consistent responses in 75\% and 68\% of cases for the \textit{independent} class, and in 69\% and 73\% of cases for \textit{before–after} relations, respectively. In contrast, smaller-scale models struggle substantially with consistency, particularly for \textit{before–after} relations; for example, InternVL3.5-8B achieves only 34\% consistency under its best-performing prompting setting. This further highlights that models struggle to reason under the reversed temporal order. Fine-tuning improves prediction consistency, with InternVL3.5-8B achieving 76\% consistency for \textit{before–after}  pairs and 73\% for \textit{independent} pairs.

\subsection{Conclusion}
We introduced MATEO, a publicly available multimodal benchmark designed to evaluate and advance the multimodal temporal reasoning and thus planning abilities of LVLMs via the Temporal Execution Order task. We provided comprehensive guidelines for extending the benchmark to new recipes, languages, and domains, enabling scalable, reproducible use for both evaluation and model development. We evaluated six open- and closed-source models and demonstrated that multimodality is critical for successfully performing the TEO task. Nevertheless, even the best-performing model achieved only modest performance, highlighting a possible explanation for the limited planning capabilities of current LVLMs. We believe that MATEO can serve as a foundation for future research and improvements in multimodal temporal reasoning and real-world planning.

\section*{Limitations}
The main limitation of this work lies in the prompting strategies. We observed that models often interpreted instructions differently, requiring minor model-specific adjustments, such as rephrasing the output format or clarifying the reasoning instructions, and preventing a fully uniform, directly comparable evaluation. Specifically, InternVL3.5-8B failed to consistently follow the required output format unless a special \textit{<assistant>} token was removed. We did not observe this issue with InternVL3.5-38b. Similarly, GPT-5.1 tended to bypass the requested chain-of-thought reasoning and produce only a final answer. To fix it, we added the instruction: "you must follow the exact reasoning steps provided in the examples before providing the final answer" at the end of the prompt. Additionally, due to resource constraints, we were unable to explore multimodal in-context learning (ICL), leaving open the question of how combining image and text examples in ICL would affect performance. Finally, our experiments were conducted only in English; thus, the findings may not generalize to other languages.

\section*{Ethical Statements}
Although we believe this work raises no direct ethical concerns, we cannot fully rule out potential dual-use implications of our findings.



\bibliography{custom}

\appendix

\section{Appendix}

\subsection{Experimental details}
\label{sec:hyperparameters}
We used the HuggingFace framework to run experiments with open-source models. For generation, we employed greedy decoding (\texttt{do\_sample=False}, \texttt{num\_beams=1}) for all open-source models to foster reproducibility. We used the following model checkpoints: Qwen2.5-VL-72b\footnote{\url{https://huggingface.co/Qwen/Qwen2.5-VL-72B-Instruct}},  Qwen2.5-VL-7b\footnote{\url{https://huggingface.co/Qwen/Qwen2.5-VL-7B-Instruct}}, InternVL3.5-8b\footnote{\url{https://huggingface.co/OpenGVLab/InternVL3_5-8B-HF}}, InternVL3.5-38b\footnote{\url{https://huggingface.co/OpenGVLab/InternVL3_5-38B-HF}}, and LLaVA-OneVision-7b\footnote{\url{https://huggingface.co/lmms-lab/llava-onevision-qwen2-7b-ov}}.

To run closed-source GPT-5.1\footnote{Model snapshot: gpt-5.1-2025-11-13} model, we used OpenAI Batch API \footnote{https://platform.openai.com/docs/guides/batch}. As with the open-source models, we used greedy decoding for reproducibility (\texttt{temperature=0}). Since we wanted the model to follow our reasoning steps (e.g. CoT experiment), we disabled the model’s default internal reasoning by setting \texttt{reasoning=\{"effort": "none"\}}.   

For fine-tuning, we used a batch size of 1, a learning rate of 1e-5, and a weight decay of 0.01. The models were fine-tuned for 1 epoch. We used Low-Rank Adaptation (LoRA) \cite{hu2021loralowrankadaptationlarge} to fine-tune the model. We set alpha to 32 and rank to 16. LoRA adapters are applied to the query, key, value, and output matrices of the attention layers. 

Experiments involving Qwen2.5-VL-72B, InternVL3.5-38B, and fine-tuning were conducted using two NVIDIA A100 GPUs with 80~GiB of memory, while all other experiments were run on a single NVIDIA A100 with 80~GiB of memory. For the experiments with GPT-5.1, we spent a total of around \$100, considering both experimentation and computing the results presented in this article. While for collecting the corpus, we spent a total of around \textsterling 800, considering both tests and the actual annotation.

The parameters used to screen the participant sample provided by Prolific were as follows:
\begin{itemize}
    \item First language: Italian
    \item Country of Birth: Italy
    \item Education: High school diploma/A-levels or above
    \item Approval rate: 95-100
    \item Number of previous submissions: 100-10000
\end{itemize}

\subsection{Additional Material}
\label{sec:materials}
\onecolumn

\begin{longtable}{c l p{10cm}}
\label{tab:prompt_settings_fullpage} \\
\textbf{Prompt Setting} & \textbf{Modality} & \textbf{Prompt} \\ \toprule
\endfirsthead
\textbf{Prompt Setting} & \textbf{Modality} & \textbf{Prompt} \\ \toprule
\endhead

\multirow{3}{*}{Baseline} 
& Image-only & Using ONLY the information in the Context, answer the following three questions in EXACTLY this format: \\
            &           & Q1: The answer is: <Yes/No/I don't know>. \\
            &           & Q2: The answer is: <Yes/No/I don't know>. \\
            &           & Q3: The answer is: <Yes/No/I don't know>. \\
            &           & Do not add anything else. Do not explain. Do not change the format. \\
            &           & Context: \\
            &           & Step A picture: [image\_A] \\
            &           & Step B picture: [image\_B] \\
            &           & Questions: \\
            &           & Q1: Must Step A be executed before Step B? \\
            &           & Q2: Must Step A be executed after Step B? \\
            &           & Q3: Can Step A and Step B be executed in parallel? \\ \cdashline{2-3}
& Text-only & Using ONLY the information in the Context ... \\
            &           & Context: \\
            &           & Step A description: [description\_A] \\
            &           & Step B description: [description\_B] \\
            &           & Questions ... \\ \cdashline{2-3}
& Image-Text & Using ONLY the information in the Context ... \\
             &           & Context: \\
             &           & Step A picture: [image\_A] \\
             &           & Step A description: [description\_A] \\
             &           & Step B picture: [image\_B] \\
             &           & Step B description: [description\_B] \\
             &           & Questions ... \\ \midrule
\multirow{3}{*}{Instructions} 
& Image-only & Your task is to determine the dependency order between two steps in a recipe. Follow these rules: \\
             &           & - \textbf{Before}: Step A must be executed before Step B if the outcome of Step A is required to complete Step B (i.e., Step B depends on Step A). \\
             &           & - \textbf{After}: Step A must be executed after Step B if the outcome of Step B is required to complete Step A (i.e., Step A depends on Step B). \\
             &           & -\textbf{Parallel}: Step A and Step B can be executed in parallel if neither step depends on the outcome of the other; therefore, their order of execution can be arbitrary. \\
             &           & Using ONLY the information in the Context ... \\
             &           & Context: ... \\
             &           & Questions: ... \\ \cdashline{2-3}
& Text-only & Your task is to determine the dependency order between two steps in a recipe. Follow these rules: \\
             &           & - Before: ... \\
             &           & - After: ... \\
             &           & - Parallel: ... \\
             &           & Ignore sequencing terms (e.g., 'first', 'then', 'lastly', and other words that may appear in the text for the natural flow of the recipe) when determining the execution order, and focus only on the action itself. \\
             &           & Using ONLY the information in the Context ... \\
             &           & Context: ... \\
             &           & Questions: ... \\ \cdashline{2-3}
& Image-Text & Your task is to determine the dependency order between two steps in a recipe. Follow these rules: \\
             &           & - Before: ... \\
             &           & - After: ... \\
             &           & - Parallel: ... \\
             &           & Ignore sequencing terms (e.g., 'first', 'then', 'lastly', and other words that may appear in the text for the natural flow of the recipe) when determining the execution order, and focus only on the action itself. Also note that the text description may include partial or full references to steps not shown in the image; in such cases, rely on the actions depicted in the image. \\
             &           & Using ONLY the information in the Context ... \\
             &           & Context: ... \\
             &           & Questions: ... \\
\midrule

\multirow{3}{*}{In-context learning}
& Image-only  & Your task is to determine the dependency order between two steps in a recipe. Follow these rules: \\
             &           & - Before: ... \\
             &           & - After: ... \\
             &           & - Parallel: ... \\
             &           & \textbf{You will be shown three examples demonstrating how to solve the task using text-based step descriptions. However, your actual input will consist of images, and your reasoning should be based on the actions depicted in those images.} \\
             &           & Examples: \\
             &           & \\
             &           & Step A description: Grate the lemon zest. \\
             &           & Step B description: Put the grated lemon zest into the strawberry sauce. \\
             &           & Questions: \\
             &           & Q1: Must Step A be executed before Step B? \\
             &           & Q2: Must Step A be executed after Step B? \\
             &           & Q3: Can Step A and Step B be executed in parallel? \\
             &           & Q1: The answer is: Yes. \\
             &           & Q2: The answer is: No. \\
             &           & Q3: The answer is: No. \\
             &           & Explanation: Step B explicitly depends on Step A - lemon zest must already be grated (Step A) before it can be put into the strawberry sauce (Step B); therefore, Step A must be executed before Step B. \\

             &           & \\
             &           & Step A description: Add the celery. \\
             &           & Step B description: Then add carrots. \\
             &           & Questions: \\
             &           & Q1: Must Step A be executed before Step B? \\
             &           & Q2: Must Step A be executed after Step B? \\
             &           & Q3: Can Step A and Step B be executed in parallel? \\
             &           & Q1: The answer is: No. \\
             &           & Q2: The answer is: No. \\
             &           & Q3: The answer is: Yes. \\
             &           & Explanation:  Both actions are independent; neither step produces something the other one requires. \\

             &           & \\
             &           & Step A description: Pour it into the cup. \\
             &           & Step B description: Measure out raspberry juice. \\
             &           & Questions: \\
             &           & Q1: Must Step A be executed before Step B? \\
             &           & Q2: Must Step A be executed after Step B? \\
             &           & Q3: Can Step A and Step B be executed in parallel? \\
             &           & Q1: The answer is: No. \\
             &           & Q2: The answer is: Yes. \\
             &           & Q3: The answer is: No. \\
             &           & Explanation:  Step A relies on the outcome of Step B - raspberry juice must be poured into the cup (Step A) after it is measured out (Step B); therefore, Step A must be executed after Step B. \\ 
             &           & Using ONLY the information in the Context ... \\
             &           & Context: ... \\
             &           & Questions: ... \\ \cdashline{2-3}

& Text-only  & Your task is to determine the dependency order between two steps in a recipe. Follow these rules: \\
             &           & - Before: ... \\
             &           & - After: ... \\
             &           & - Parallel: ... \\
             &           & Ignore sequencing terms ... \\
             &           & \textbf{You will be shown three examples demonstrating how to solve the task using text-based step descriptions.} \\
             &           & Examples: ... \\
             &           & Using ONLY the information in the Context ... \\
             &           & Context: ... \\
             &           & Questions: ... \\  \cdashline{2-3}
& Image-Text & Your task is to determine the dependency order between two steps in a recipe. Follow these rules: \\
             &           & - Before: ... \\
             &           & - After: ... \\
             &           & - Parallel: ... \\
             &           & Ignore sequencing terms ... \\
             &           & \textbf{You will be shown three examples demonstrating how to solve the task using text-based step descriptions. However, your actual input will consist of both images and text descriptions, and your reasoning should be based on both actions shown in the images and the accompanying textual descriptions.} \\
             &           & Examples: ... \\
             &           & Using ONLY the information in the Context ... \\
             &           & Context: ... \\
             &           & Questions: ... \\  \midrule

\multirow{3}{*}{CoT} 
& Image-only & Your task is to determine the dependency order between two steps in a recipe. \textbf{You must choose from: Before, After, or Parallel}. Follow these rules: \\
             &           & - Before: ... \\
             &           & - After: ... \\
             &           & - Parallel: ... \\
             &           & \textbf{You will be shown three examples demonstrating how to solve the task using text-based step descriptions. However, your actual input will consist of images, and your reasoning should be based on the actions depicted in those images. You must follow the reasoning steps shown in the examples before answering.} \\
             &           & \\
             &           & Examples: \\
             &           & \\
             &           & Step A description: Grate the lemon zest. \\
             &           & Step B description: Put the grated lemon zest into the strawberry sauce. \\
             &           & Step A produces: Grated lemon zest. \\
             &           & Step B produces: Lemon zest inside the strawberry sauce. \\
             &           & Step A requires: A lemon. \\
             &           & Step B requires: Lemon zest that has been grated. \\
             &           & Dependency analysis: Step B explicitly depends on Step A - lemon zest must already be grated (Step A) before it can be put into the strawberry sauce (Step B); therefore, Step A must be executed before Step B. \\
             &           & The answer is: Before. \\
             &           & \\
             &           & Step A description: Add the celery. \\
             &           & Step B description: Then add carrots. \\
             &           & Step A produces: A component with the celery added. \\
             &           & Step B produces: A component with the carrots added. \\
             &           & Step A requires: The celery. \\
             &           & Step B requires: The carrots. \\
             &           & Dependency analysis: Each step adds a separate ingredient, and neither depends on the other, so they can occur in any order. \\
             &           & The answer is: Parallel. \\
             &           & \\
             &           & Step A description: Pour it into the cup. \\
             &           & Step B description: Measure out raspberry juice. \\
             &           & Step A produces: The cup with raspberry juice poured in it. \\
             &           & Step B produces: Raspberry juice that was measured out. \\
             &           & Step A requires: Raspberry juice that was measured out. \\
             &           & Step B requires: Raspberry juice. \\
             &           & Dependency analysis: Step A relies on the outcome of Step B - raspberry juice must be poured into the cup (Step A) after it is measured out (Step B); therefore, Step A must be executed after Step B. \\
             &           & The answer is: After. \\
             &           & \\
             &           & Step A picture: ... \\ \cdashline{2-3}

& Text-only  & Your task is to determine the dependency order between two steps in a recipe. \textbf{You must choose from: Before, After, or Parallel}. Follow these rules: \\
             &           & - Before: ... \\
             &           & - After: ... \\
             &           & - Parallel: ... \\
             &           & Ignore sequencing terms ... \\
             &           & \textbf{You will be shown three examples demonstrating how to solve the task using text-based step descriptions. You must follow the reasoning steps shown in the examples before answering.} \\
             &           & \\
             &           & Examples: ... \\
             &           & \\
             &           & Step A description: ... \\ \cdashline{2-3}

& Image-Text & Your task is to determine the dependency order between two steps in a recipe. \textbf{You must choose from: Before, After, or Parallel}. Follow these rules: \\
             &           & - Before: ... \\
             &           & - After: ... \\
             &           & - Parallel: ... \\
             &           & \textbf{You will be shown three examples demonstrating how to solve the task using text-based step descriptions. However, your actual input will consist of both images and text descriptions, and your reasoning should be based on both actions shown in the images and the accompanying textual descriptions. You must follow the reasoning steps shown in the examples before answering.} \\
             &           & \\
             &           & Examples: ... \\
             &           & \\
             &           & Step A picture: ... \\
             &           & Step A description: ... \\ \midrule

\multirow{3}{*}{Self-reflection} 
& Image-only & Your task is to determine the dependency order between two steps in a recipe. \textbf{You must choose from: Before, After, or Parallel}. Follow these rules: \\
             &           & - Before: ... \\
             &           & - After: ... \\
             &           & - Parallel: ... \\
             &           & You will be shown three examples demonstrating how to solve the task using text-based step descriptions. However, your actual input will consist of images, and your reasoning should be based on the actions depicted in those images. You must follow the reasoning steps shown in the examples before answering. \textbf{After you answer the question, review your reasoning and check whether your answer logically follows from the context and dependencies you identified. After self-reflection, provide your final answer, confirming or correcting your initial choice.} \\
             &           & \\
             &           & Examples: \\
             &           & \\
             &           & Step A description: Grate the lemon zest. \\
             &           & Step B description: Put the grated lemon zest into the strawberry sauce. \\
             &           & Step A produces: Grated lemon zest. \\
             &           & Step B produces: Lemon zest inside the strawberry sauce. \\
             &           & Step A requires: A lemon. \\
             &           & Step B requires: Lemon zest that has been grated. \\
             &           & Dependency analysis: Step B explicitly depends on Step A - lemon zest must already be grated (Step A) before it can be put into the strawberry sauce (Step B); therefore, Step A must be executed before Step B. \\
             &           & The answer is: Before. \\
             &           & \textbf{Reflection: <your\_reflection>} \\
             &           & \textbf{The final answer: <final\_answer>} \\
             &           & \\
             &           & ... \\
             &           & \\
             &           & Step A picture: ... \\ \cdashline{2-3}

& Text-only  & Your task is to determine the dependency order between two steps in a recipe. \textbf{You must choose from: Before, After, or Parallel}. Follow these rules: \\
             &           & - Before: ... \\
             &           & - After: ... \\
             &           & - Parallel: ... \\
             &           & Ignore sequencing terms ... \\
             &           & You will be shown three examples demonstrating how to solve the task using text-based step descriptions. You must follow the reasoning steps shown in the examples before answering. \textbf{After you answer the question, review your reasoning and check whether your answer logically follows from the context and dependencies you identified. After self-reflection, provide your final answer, confirming or correcting your initial choice.} \\
             &           & \\
             &           & Examples: ... \\
             &           & \\
             &           & Step A description: ... \\ \cdashline{2-3}

& Image-Text & Your task is to determine the dependency order between two steps in a recipe. \textbf{You must choose from: Before, After, or Parallel}. Follow these rules: \\
             &           & - Before: ... \\
             &           & - After: ... \\
             &           & - Parallel: ... \\
             &           & \textbf{You will be shown three examples demonstrating how to solve the task using text-based step descriptions. However, your actual input will consist of both images and text descriptions, and your reasoning should be based on both actions shown in the images and the accompanying textual descriptions. You must follow the reasoning steps shown in the examples before answering.} \\
             &           & \\
             &           & Examples: ... \\
             &           & \\
             &           & Step A picture: ... \\
             &           & Step A description: ... \\ \midrule

\caption{Prompt schema used for models evaluation under image-only, text-only, and image–text modalities. Repeated or previously described components are omitted for brevity.}
\label{table:prompt_schema}
\end{longtable}

\newpage
\begin{longtable}{l p{12cm}}
\label{tab:prompt_settings_fullpage} \\
\textbf{Modality} & \textbf{Prompt} \\ \toprule
\endfirsthead
\endhead

\multirow{2}{*}{Image-only} 
& <User> Your task is to determine the dependency order between two steps in a recipe. Follow these rules: \\
                        & - \textbf{Before}: Step A must be executed before Step B if the outcome of Step A is required to complete Step B (i.e., Step B depends on Step A). \\
                        & - \textbf{After}: Step A must be executed after Step B if the outcome of Step B is required to complete Step A (i.e., Step A depends on Step B). \\
                       & -\textbf{Parallel}: Step A and Step B can be executed in parallel if neither step depends on the outcome of the other; therefore, their order of execution can be arbitrary. \\
                       & Step A picture: [image\_A] \\
                       & Step B picture: [image\_B] \\
                       & Questions: \\
                       & Q1: Must Step A be executed before Step B? \\
                       & Q2: Must Step A be executed after Step B? \\
                       & Q3: Can Step A and Step B be executed in parallel? \\ 
                       & <Assistant> \\
                       & Q1: The answer is: Yes. \\
                       & Q2: The answer is: No. \\
                       & Q3: The answer is: No. \\ 
\midrule
\multirow{2}{*}{Text-only} 
& <User> Your task is to determine the dependency order between two steps in a recipe. Follow these rules: \\
                        & - Before: ... \\
                       & Ignore sequencing terms (e.g., 'first', 'then', 'lastly', and other words that may appear in the text for the natural flow of the recipe) when determining the execution order, and focus only on the action itself. \\
                       & Step A description: [description\_A] \\
                       & Step B description: [description\_B] \\
                       & Questions: \\
                       & Q1: Must Step A be executed before Step B? \\
                       & Q2: Must Step A be executed after Step B? \\
                       & Q3: Can Step A and Step B be executed in parallel? \\ 
                       & <Assistant> \\
                       & Q1: The answer is: No. \\
                       & Q2: The answer is: No. \\
                       & Q3: The answer is: Yes. \\ 
\midrule
\multirow{2}{*}{Image-Text} 
& <User> Your task is to determine the dependency order between two steps in a recipe. Follow these rules: \\
                        & - Before: ... \\
                       & Ignore sequencing terms (e.g., 'first', 'then', 'lastly', and other words that may appear in the text for the natural flow of the recipe) when determining the execution order, and focus only on the action itself.  Also note that the text description may include partial or full references to steps not shown in the image; in such cases, rely on the actions depicted in the image. \\
                       & Step A description: [description\_A] \\
                       & Step B description: [description\_B] \\
                       & Questions: \\
                       & Q1: Must Step A be executed before Step B? \\
                       & Q2: Must Step A be executed after Step B? \\
                       & Q3: Can Step A and Step B be executed in parallel? \\ 
                       & <Assistant> \\
                       & Q1: The answer is: No. \\
                       & Q2: The answer is: Yes. \\
                       & Q3: The answer is: No. \\
\caption{Example of prompt–response pairs used for fine-tuning across image-only, text-only, and image–text modalities. Each example includes instructions, input structure, and expected binary outputs. The \textless User\textgreater{} and \textless Assistant\textgreater{} tokens are model-specific and are included here solely for illustrative purposes.}
\label{tab:fine_tuning_ex}
\end{longtable}

\newpage

\begin{figure}
    \centering
    \includegraphics[width=\textwidth]{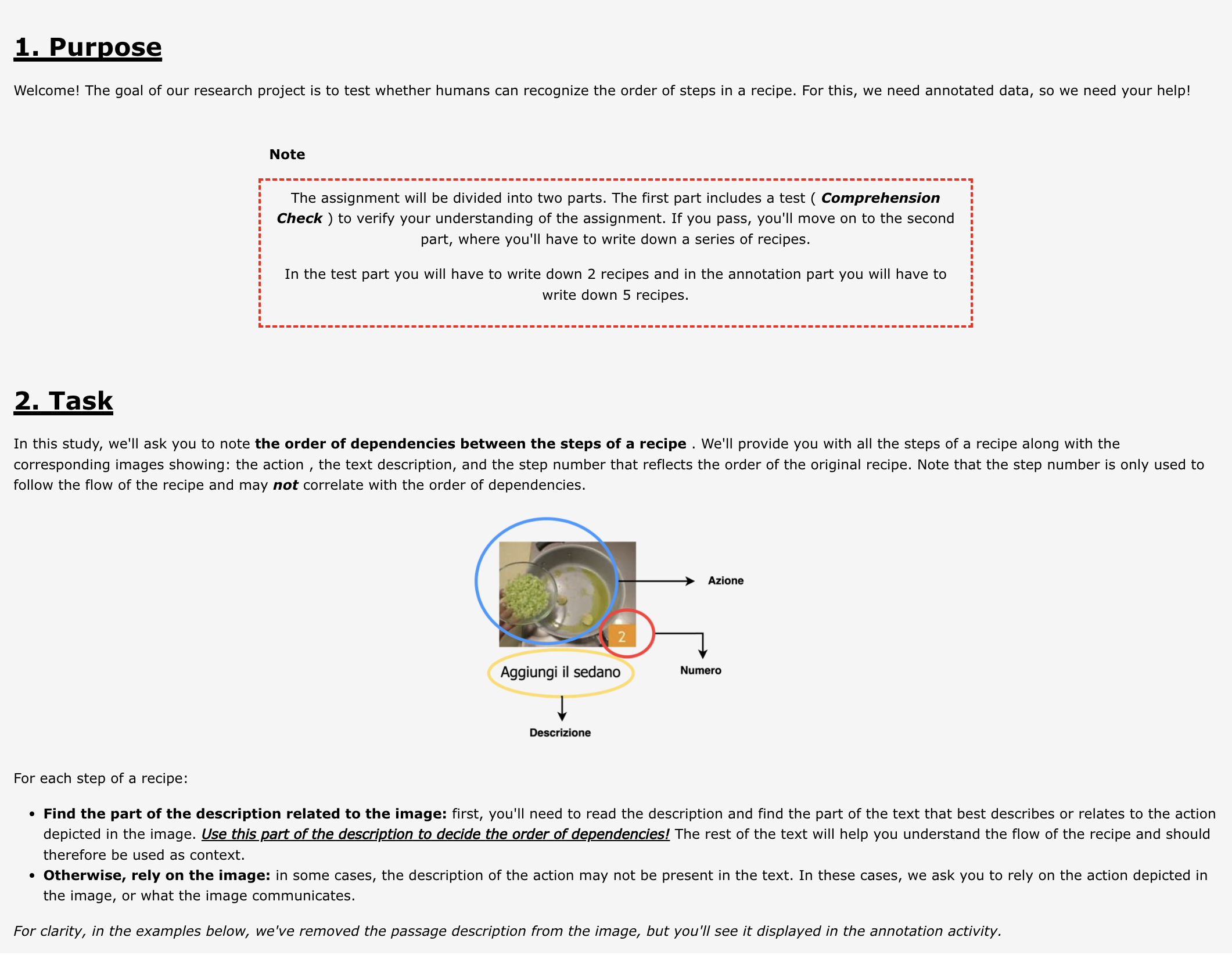}
    \caption{This figure shows the first two points of the annotation guidelines, that present the purpose and the general description of the task.}
    \label{fig:ann1}
\end{figure}

\begin{figure}
    \centering
    \includegraphics[width=\textwidth]{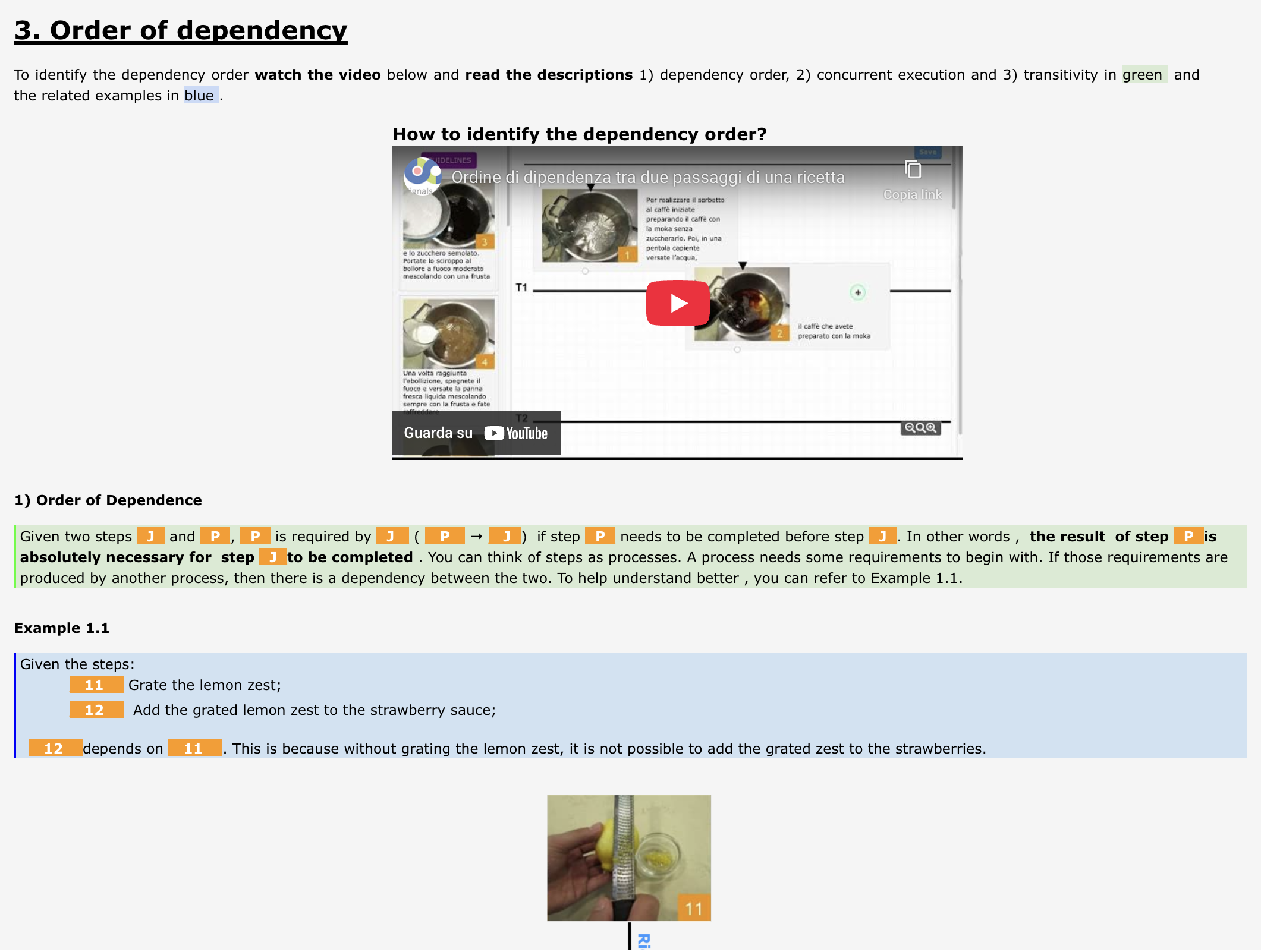}
    \caption{This section of the annotation guidelines offers annotators the option to watch a demonstration video that walks through the task and explains how to identify dependencies between steps. Following the video, annotators are provided with a textual description that formally defines each dependency, accompanied by illustrative examples. }
    \label{fig:ann2}
\end{figure}

\begin{figure}
    \centering
    \includegraphics[width=\textwidth]{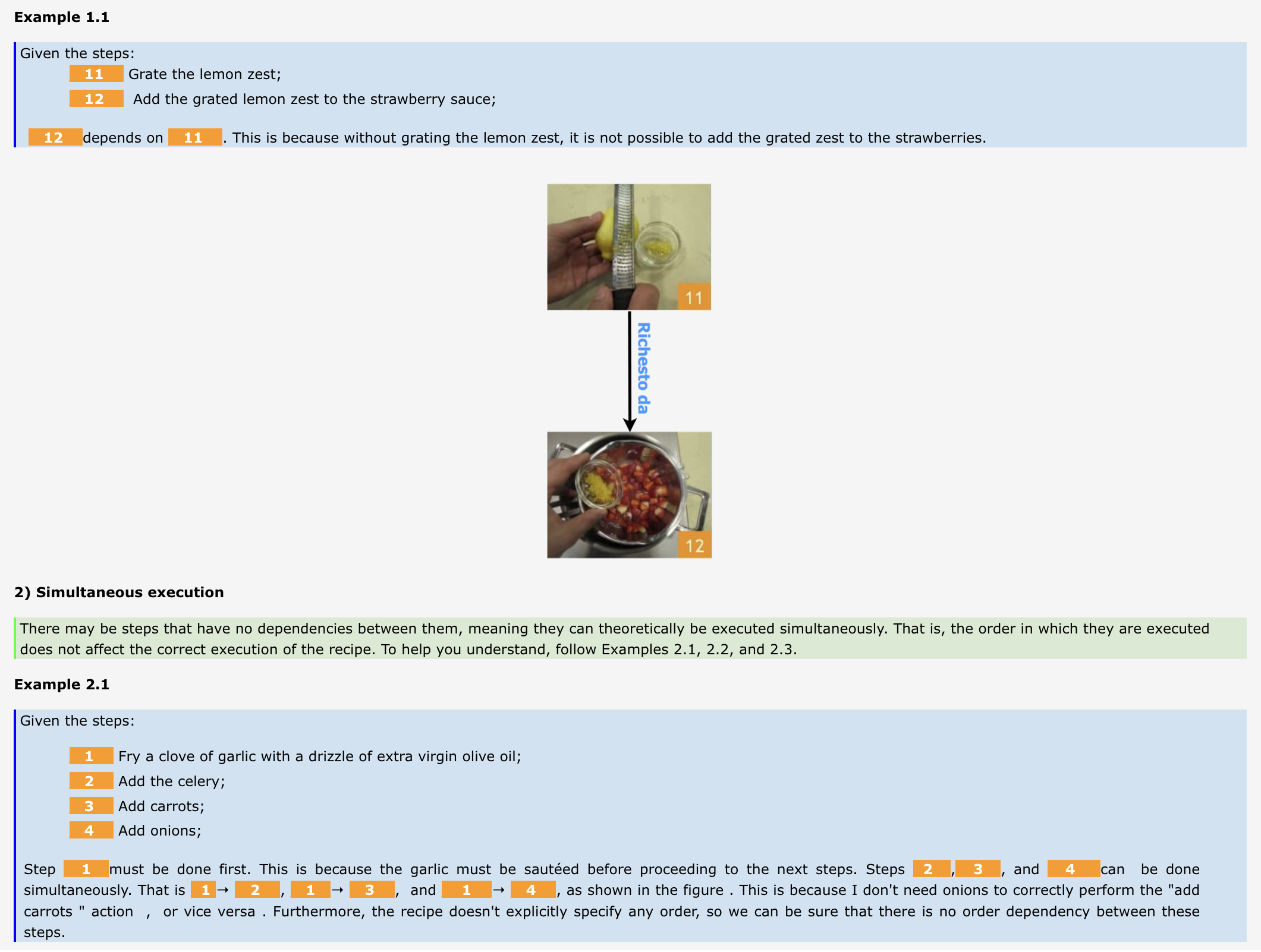}
    \caption{This figure continues from the previous one, illustrating a case in which a step depends on another (Example 1.1), and presenting the formal definition of independent steps }
    \label{fig:ann3}
\end{figure}

\begin{figure}
    \centering
    \includegraphics[width=\textwidth]{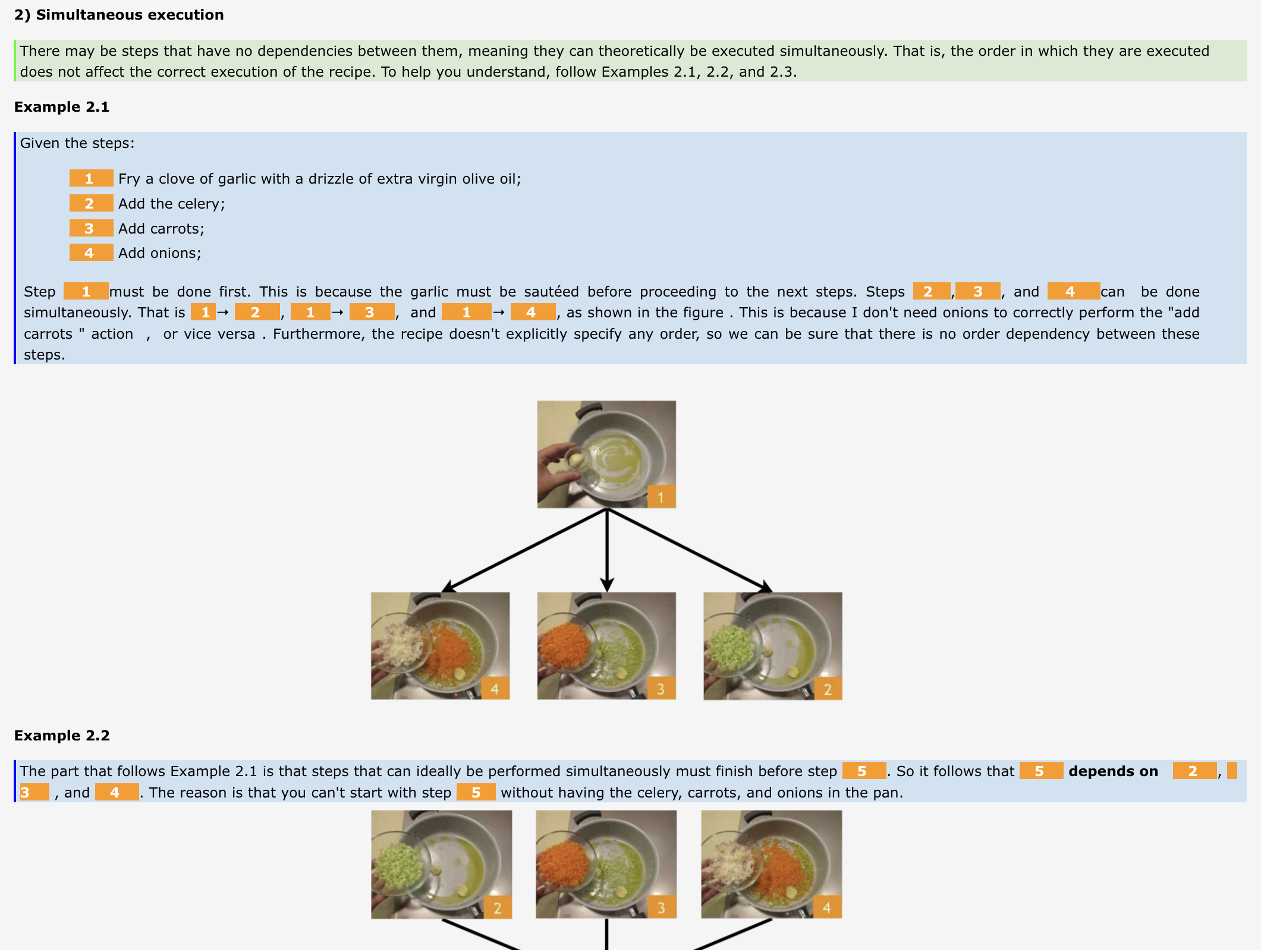}
    \caption{This figure shows an example given to illustrate the case in which steps are independent. }
    \label{fig:ann4}
\end{figure}

\begin{figure}
    \centering
    \includegraphics[width=\textwidth]{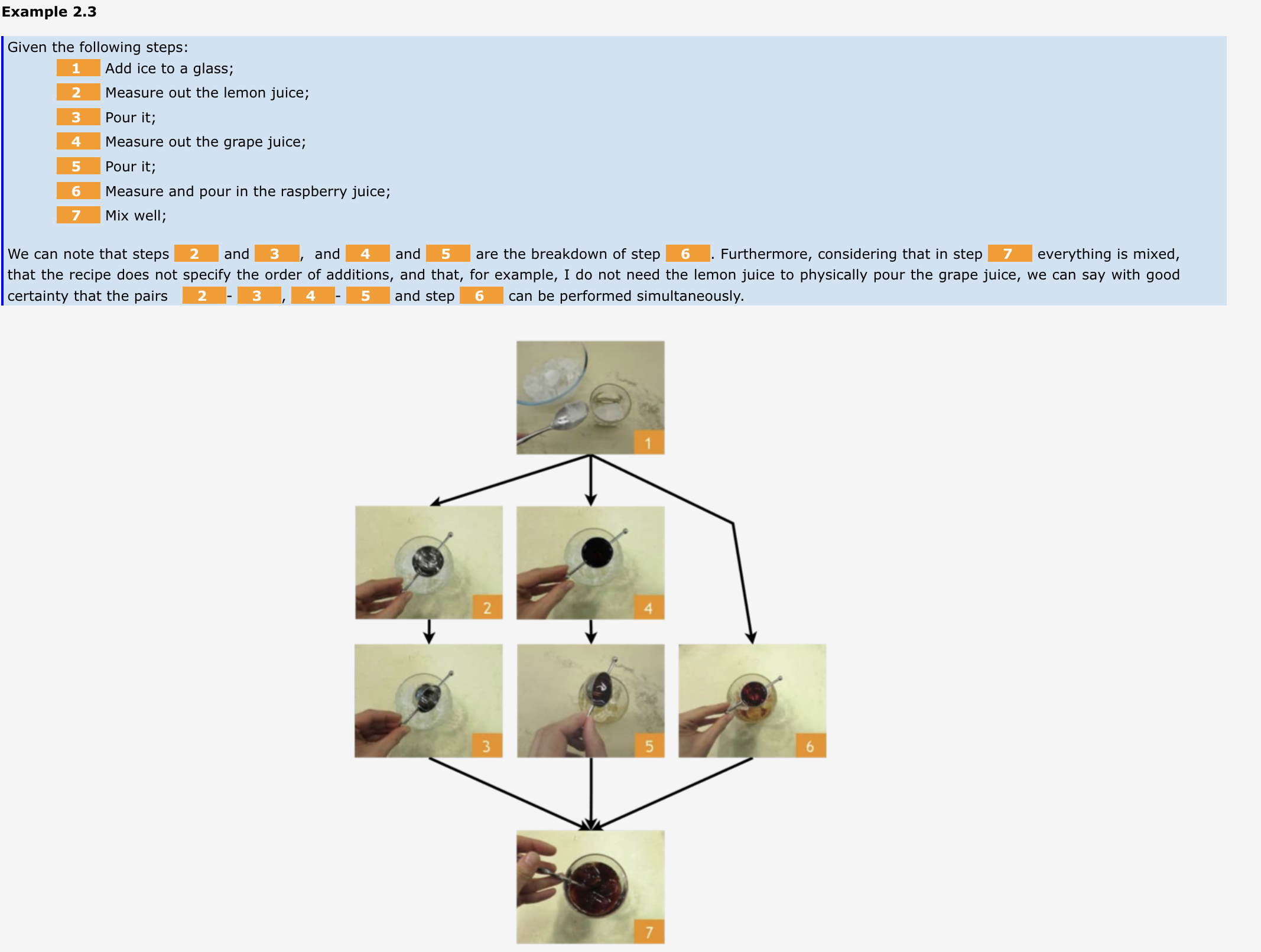}
    \caption{Example illustrating a fully annotated DAG for a recipe, with all relations clearly explained.}
    \label{fig:ann5}
\end{figure}

\begin{figure}
    \centering
    \includegraphics[width=\textwidth]{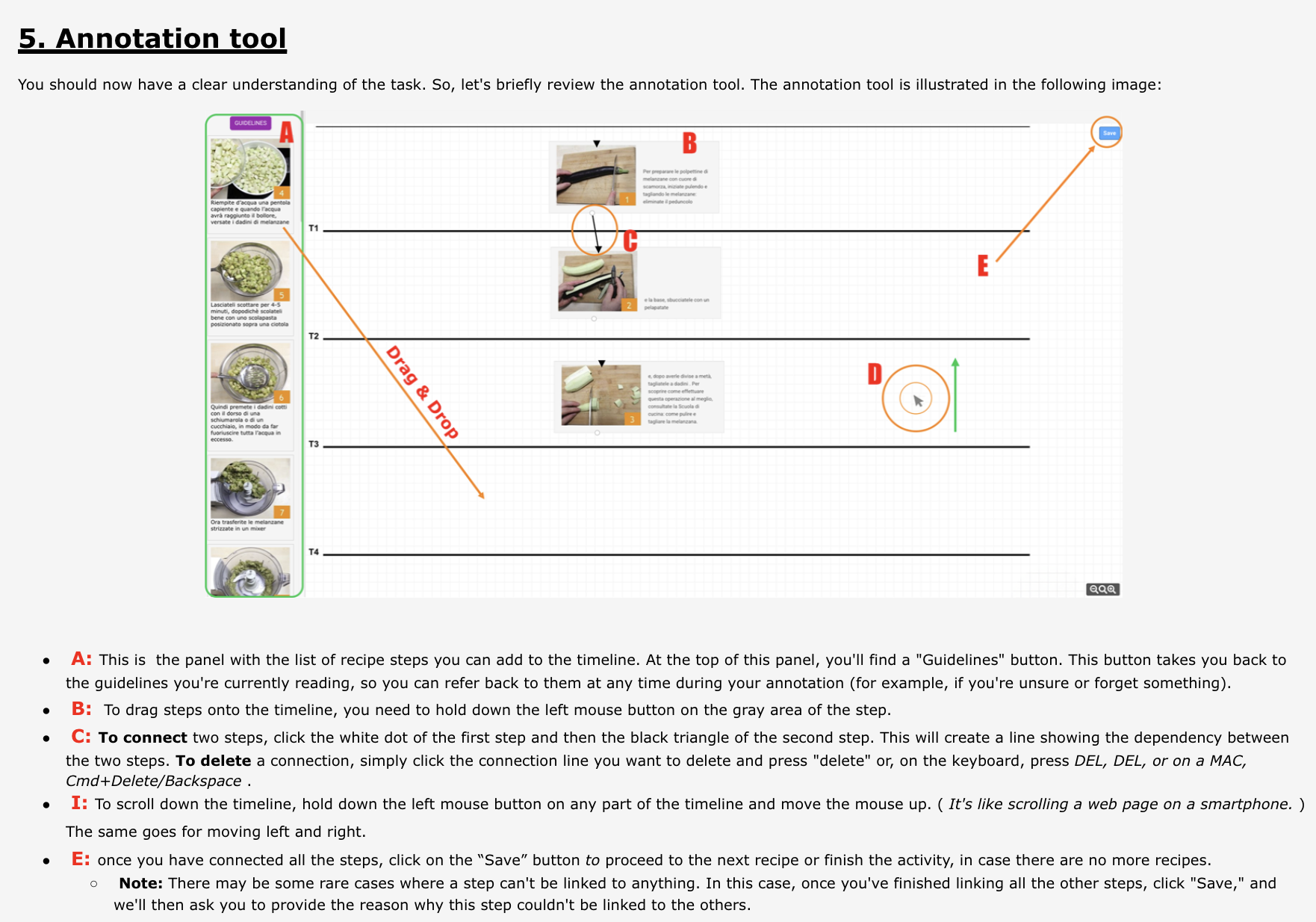}
    \caption{Illustration of the annotation tool as presented in the annotation guidelines. Highlighted regions (marked in red letters) indicate key functionalities, including dragging steps, adding or deleting connections, and saving the annotations. }
    \label{fig:annotation_platform}
\end{figure}

\begin{figure}
    \centering
    \includegraphics[width=\textwidth]{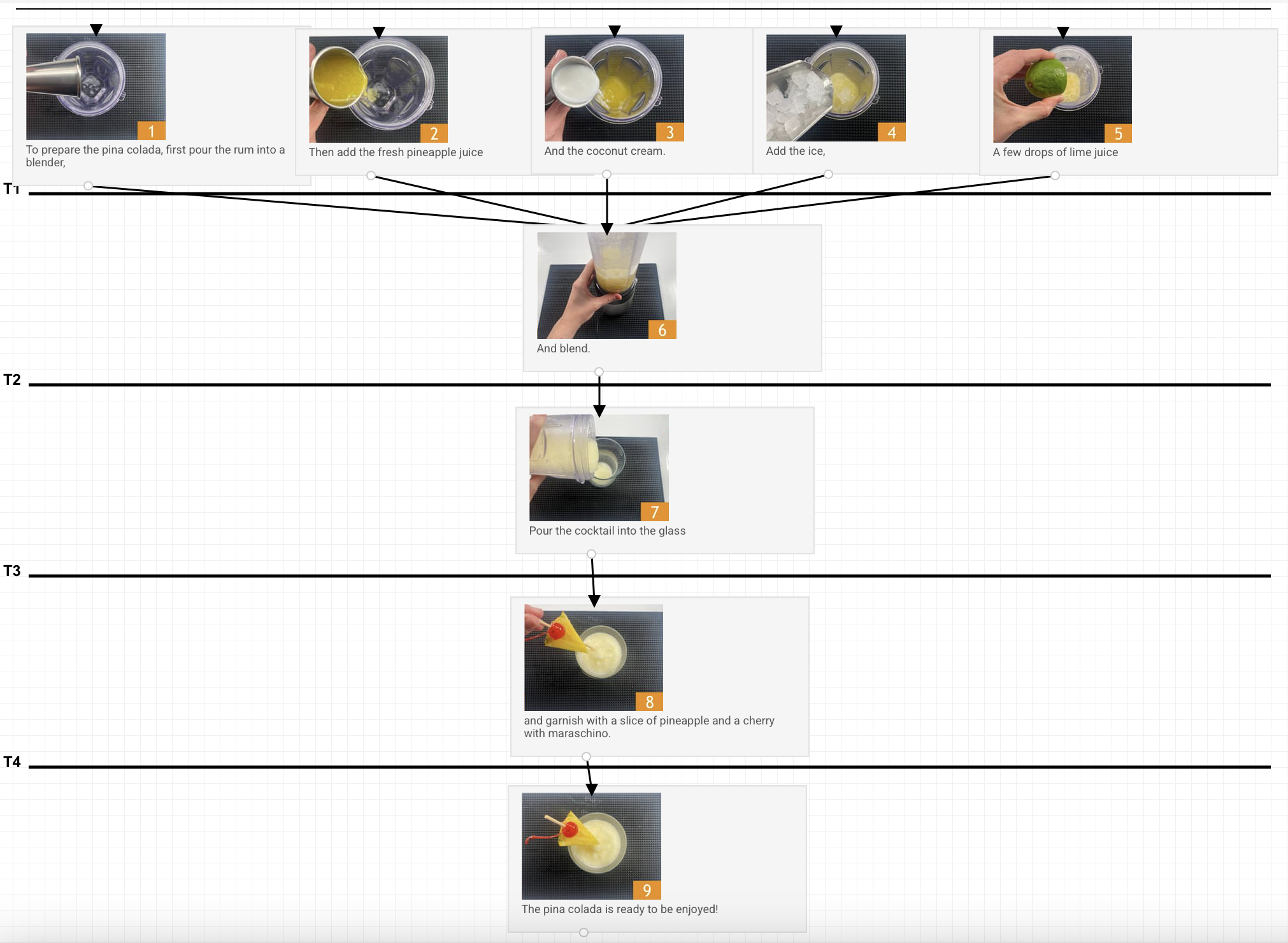}
    \caption{Example of a DAG annotated by a crowd-worker. Steps 1, 2, 3, 4, and 5 can be executed in any order (or simultaneously), thus there is no directed edge between them. }
    \label{fig:dag_ex}
\end{figure}
\subsection{AI tool usage}
During the development phase, we used ChatGPT and Gemini to identify and resolve implementation bugs. We also leveraged these models and Grammarly to suggest rephrasing that improves the fluency and readability of the manuscript. Additionally, we have experimented with using Google Scholar Labs to expand and refine our literature review.

\subsection{Licences of the corpus and models used}
All models used in our experiments have been utilised in accordance with their respective licenses. We plan to license the corpus with the Creative Commons 4.0 CC BY-NC-SA 4.0 (Attribution-NonCommercial-ShareAlike). The intended use of this corpus will be only for scientific research purposes.

\end{document}